\documentclass[letterpaper, 10 pt, conference]{ieeeconf}  

\IEEEoverridecommandlockouts                              

\usepackage{graphicx}
\usepackage{amsmath} 
\usepackage{amssymb}  

\title{\LARGE \bf
Learning Content-Aware Multi-Modal Joint Input Pruning via Birds'-Eye-View Representation
}

\author{
    Yuxin Li$^{1,2}$ 
    Yiheng Li$^{1}$ 
    Xulei Yang$^{3}$ 
    Mengying Yu$^{2}$
    Zihang Huang$^{2}$ 
    Xiaojun Wu$^{2}$ 
    Chai Kiat Yeo$^{1}$ 
    \thanks{*This work is sponsored by Desay SV Singapore}
    \thanks{1 School of Computer Science and Engineering, Nanyang Technological University, Singapore
        {\tt\small  yuxin004@e.ntu.edu.sg;}
                    }%
    \thanks{2 Desay SV Automotive Singapore}%
    \thanks{3 Institute for Infocomm Research (I2R), Agency for Science, Technology and Research (A*STAR), Singapore}%
}
\begin{document}

\maketitle
\thispagestyle{empty}
\pagestyle{empty}

\begin{abstract}
In the landscape of autonomous driving, Bird's-Eye-View (BEV) representation has recently garnered substantial attention, serving as a transformative framework for the fusion of multi-modal sensor inputs. The BEV paradigm effectively shifts the sensor fusion challenge from a rule-based methodology to a data-centric approach, thereby facilitating more nuanced feature extraction from an array of heterogeneous sensors. Notwithstanding its evident merits, the computational overhead associated with BEV-based techniques often mandates high-capacity hardware infrastructure, thus posing challenges for practical, real-world implementations. To mitigate this limitation, we introduce a novel content-aware multi-modal joint input pruning technique. Our method leverages BEV as a shared anchor to algorithmically identify and eliminate non-essential sensor regions prior to their introduction into the perception model's backbone. We validate the efficacy of our approach through extensive experiments on the NuScenes dataset, demonstrating substantial computational efficiency without sacrificing perception accuracy. To the best of our knowledge, this work represents the first attempt to alleviate the computational burden from the input pruning point.
\end{abstract}

\section{Introduction}
In recent years, Bird's-Eye-View (BEV) representation has garnered considerable interest as a potent mechanism for sensor fusion at the feature layer within the domain of autonomous driving. By constructing a unified top-down representational space derived from heterogeneous sensors such as cameras and LiDARs, BEV-based methodologies \cite{li2022bevformer, 2022beverse, xie2022m2bev, huang2022bevdet4d, li2022bevdepth} have exhibited a performance edge over conventional early and late fusion techniques, a superiority substantiated by their consistent high rankings on numerous public evaluation leaderboards. The intrinsic spatial coherence provided by BEV frameworks allows for a more nuanced and effective integration of diverse sensor modalities. Moreover, the BEV paradigm affords enhanced efficiency in the extraction of feature representations from voluminous datasets, thereby elevating the performance ceiling for an array of downstream tasks. As such, the implementation of multi-modal sensor fusion approaches within the BEV framework presents itself as a highly promising trajectory for advancing the perceptual efficacy of autonomous vehicular systems.

\begin{figure}[t]
\begin{center}
\includegraphics[width=\linewidth]{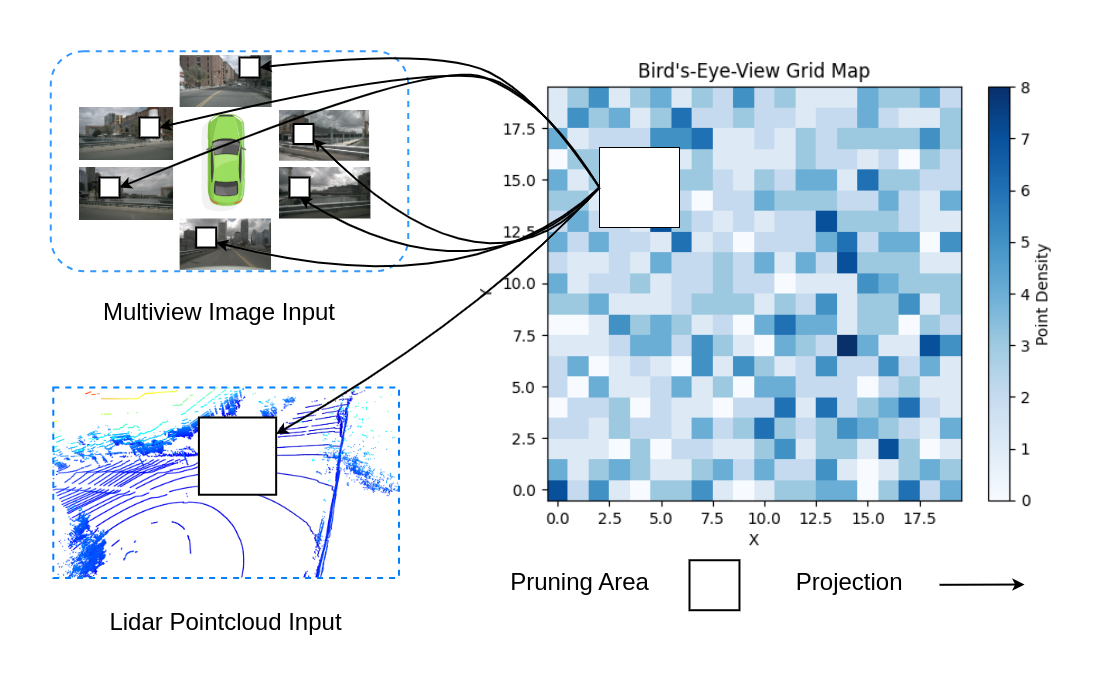}
\end{center}
\caption{Illustration of Multi-Modal Projection Relationship. The areas to be removed in the sensor space can be inferred from the anchor position in the BEV space via the intrinsic and extrinsic parameters of the sensors}
\label{fig:mutual_projection}
\end{figure}

Nonetheless, the broader deployment of these advanced methodologies is considerably hampered by their extensive computational demands, frequently exceeding the processing capabilities of conventional on-board computing systems in autonomous vehicles. These state-of-the-art (SOTA) methods typically employ a diverse range of sensor inputs, including multi-view cameras and multiple LiDAR systems capable of scanning millions of points. Initially, these inputs are subjected to complex feature extraction via specialized backbone models, followed by a projection and fusion process in the BEV space, which subsequently serves as the common representational layer for a multitude of downstream tasks. It is noteworthy that the computational bottleneck is primarily attributable to the backbone models employed in multi-modal BEV-oriented frameworks. While the incorporation of such diverse sensor inputs undoubtedly enriches the feature extraction stage, it is crucial to recognize that not all regions within the multi-modal sensor space contribute equally to the performance or accuracy of downstream applications. Consequently, a pressing research challenge lies in the development of efficient algorithms that can judiciously harness the multi-faceted spatial and temporal information available from diverse sensors while simultaneously minimizing the computational burden intrinsic to such methodologies.

\begin{figure*}[ht]
\begin{center}
\includegraphics[scale=0.85, width=\linewidth]{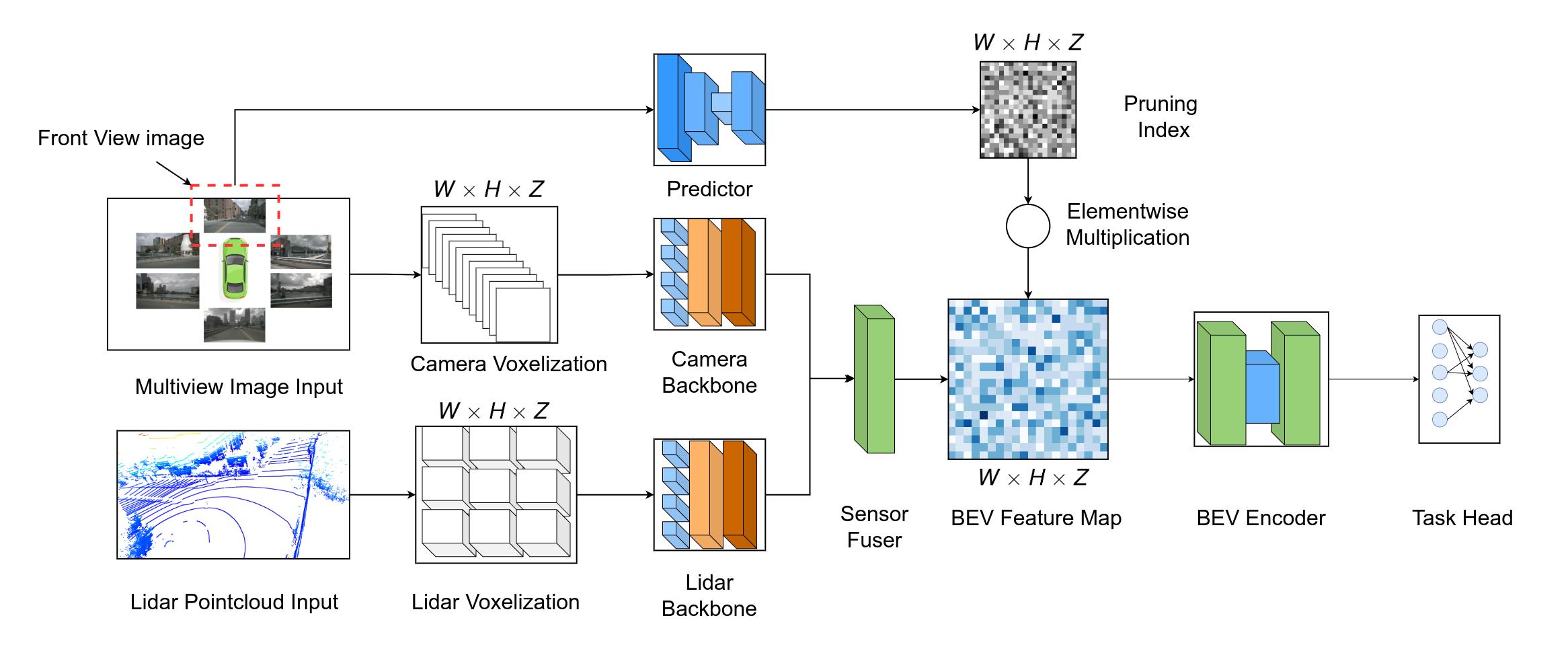}
\end{center}
\caption{Overall Training Architecture. This system combines a multi-modal perception model and a pruning index predictor. During the training phase, the predictor is trained along with the perception model in an end-to-end manner. The size of the input voxel grids, pruning index and BEV feature map are set to be identical to serve the purpose of joint pruning of multi-modal inputs.}
\label{fig:training_arch}
\end{figure*}

Inspired by the selective attention mechanisms inherent to human visual perception, we introduce a novel data-driven, content-aware, multi-modal joint input pruning approach to optimize the computational efficiency of BEV-based sensor fusion methods. Conventional Region-of-Interest (ROI) pruning techniques often resort to heuristic and geometry-based selectors that can result in imprecise object boundaries and are sensitive to sensor type disparities, thereby adversely affecting the perception task's performance. In contrast, as shown in Fig. \ref{fig:mutual_projection}, our proposed methodology prioritizes the processing of task-critical regions by jointly eliminating computationally expensive yet less important areas captured by the heterogeneous sensors, such as sky, building facades and distant static objects, without compromising the vehicle's operational safety or effectiveness. This is achieved through a spatially-aligned confidence-driven prediction module, which partitions multi-modal inputs, including camera and LiDAR data, into a grid of voxel cells. Utilizing the BEV feature map as a unified anchor representation, these cells are forward processed through sparse feature extraction backbones, and their relevance is evaluated end-to-end during the concurrent training of downstream perception tasks, and then backward projected to the source regions of the less important inputs during testing. Preliminary results from ablative experiments indicate that our approach retains comparable performance to SOTA methods even when pruning up to 50\% of the input data, thus promising a more efficient perception pipeline for autonomous vehicles. 

We summarise our contributions as follow: 
\begin{enumerate}
\item We propose a novel multi-modal joint input pruning method, resulting in a substantial 35\% reduction in model complexity. To the best of our knowledge, this is the first work to investigate efficient perception techniques from the perspective of input pruning.
\item Through rigorous experiments, we ascertain the efficacy of our pruning methodology on tasks such as 3D detection and map segmentation. Our findings underscore that our approach can eliminate more than 50\% of the redundant raw inputs while still attaining a performance that is on par with SOTA methods.
\end{enumerate}

\begin{figure*}[!ht]
\begin{center}
\includegraphics[scale=0.85, width=\linewidth, height=10cm]{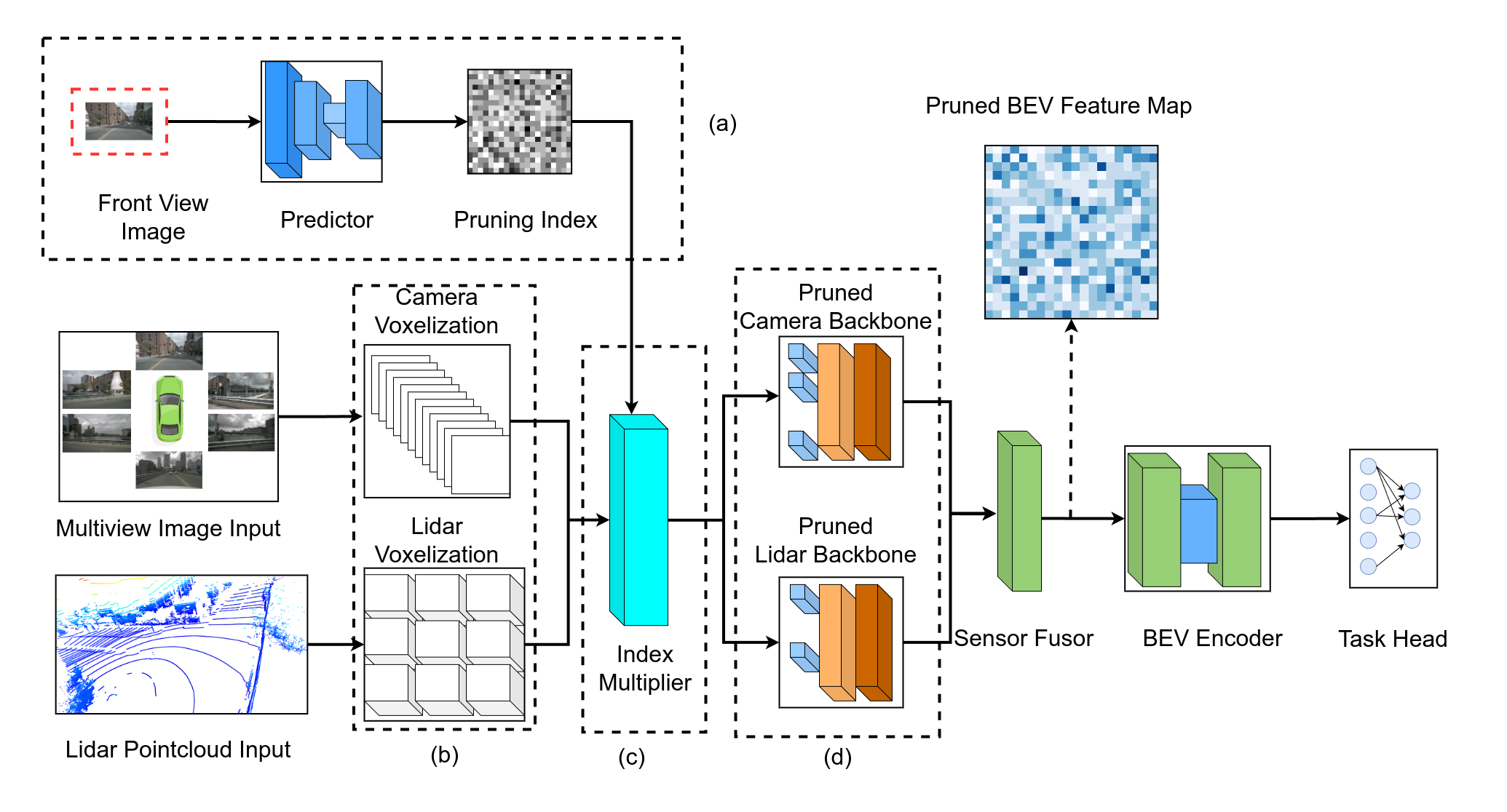}
\end{center}
\caption{Overall Inference Architecture. In the testing phase, the predictor initially generates a pruning mask, as depicted in module (a). Subsequently, it employs the index multiplier, illustrated in module (c), to align the pruning mask with the voxelized input, shown in module (b), before proceeding to the sparse backbone in module (d).}
\label{fig:testing_arch}
\end{figure*}

\section{Related Work}
\subsection{Perception in BEV Space}
Perception in the BEV space remains crucial in the realm of autonomous driving, offering a comprehensive understanding of the surrounding environment. Multiple strategies harness BEV representation to fuse LiDAR and camera inputs to foster reliable perception results for autonomous vehicles.

\textbf{3D Detection in BEV} 3D object detection serves as a foundational task in autonomous driving, facilitating navigation and decision-making. A host of techniques now optimize 3D detection using BEV space features. Contemporary vision-based methods \cite{li2022bevformer, huang2021bevdet, huang2022bevdet4d, li2022bevdepth}, integrating multi-view camera data and depth supervision, have improved by 30 points over non-BEV techniques. Additionally, multi-modal BEV methods \cite{liu2022bevfusion, harley2022simplebev, futr3d2023chen, cmt2023yan} have achieved notable performance boost, with a 10-point improvement on the NuScenes leaderboard in recent years.

\textbf{Segmentation in BEV} In map segmentation, a multitude of approaches \cite{2021hdmapnet, 2022bevsegformer, li2022vectormapnet, maptr2023liao} have leveraged the BEV space for refined scene understanding. HDMapNet \cite{2021hdmapnet} predicts vectorized BEV elements via semantic and instance-level learning, while BEVSegFormer \cite{2022bevsegformer} employs a transformer-based setup for real-time segmentation, demonstrating resilience to camera noise.

\subsection{Model Pruning}
Within the broader field of model compression, model pruning specifically addresses the challenge of streamlining neural network architectures for deployment in resource-limited environments. This approach seeks to optimize the efficiency of deep learning models by eliminating redundant or less significant nodes and branches.

The process of model pruning is methodically applied during the model's training and testing phase, where unnecessary network components are identified and removed through various strategies. These include assessing the magnitude of weights to pinpoint negligible connections \cite{weightpruning2017zhuang}, calibration based on the distribution of activations within a batch \cite{filterpruning2017hao}, implementing structural adjustments to slim down the architecture \cite{depgraph23gongfan}, and leveraging learning-based techniques to determine the expendability of certain neurons \cite{prune_astar2021, prune_kong2022spvit, prune_ltp2021, prune_toconv2021}.

The essence of model pruning lies in its capacity to reduce the complexity of neural networks without substantially compromising their performance. This balance is crucial for enabling advanced machine learning applications on devices with stringent computational constraints, highlighting the technique's significance in the ongoing evolution of model compression methodologies.

\section{Methodology}
\subsection{Problem Formulation}
The primary goal is to reduce computational overhead in processing sensor inputs by removing redundant regions. This is achieved by designing a predictor that identifies unimportant regions in a BEV representation formed by fusing multi-modal sensor inputs (Camera and LiDAR).

Formally, given a set of raw sensor inputs from both camera \( \mathcal{C} \) and LiDAR \( \mathcal{L} \), our goal is to find a function \( f \) that maps these inputs to a reduced set of inputs by pruning irrelevant or redundant regions, such that:
\begin{equation}
f(\mathcal{C}, \mathcal{L}) \rightarrow \mathcal{C}', \mathcal{L}'
\end{equation}
where \( \mathcal{C}' \) and \( \mathcal{L}' \) are the pruned inputs, with the dimensions reduced for faster inference without significant loss of pertinent information.

\textbf{Bird's-Eye-View Representation}
The BEV representation serves as an intermediary layer that combines features extracted from both camera and LiDAR inputs. For each sensor, we employ a backbone model to extract high-dimensional features. These features are then mapped into a common BEV space of dimensions \( W \times H \times Z \), where each cell in this space corresponds to a physical occupancy in the 3D world. 

\begin{equation}
\text{BEV}(\mathcal{C}, \mathcal{L}) = \text{BEV}(\mathcal{B}(\mathcal{C}), \mathcal{B}(\mathcal{L}))
\end{equation}
where \( \mathcal{B} \) is the backbone model for feature extraction.

\textbf{Predictor Representation}
We let the predictor \( P \) be trained jointly with a multi-modal perception model to identify and prune unimportant regions in the BEV representation. The predictor assigns an importance score \( s_i \) to each cell in the BEV representation.

\begin{equation}
s_i = P(\text{cell}_i)
\end{equation}
where:
\begin{equation}
s_i \in [0,1]
\end{equation}

A threshold \( \theta \) is defined to categorize the cells. If \( s_i < \theta \), the corresponding region in the raw sensor inputs is deemed redundant and pruned using the intrinsic and extrinsic parameters of the sensors. This results in the reduced inputs \( \mathcal{C}' \) and \( \mathcal{L}' \).

\subsection{Model Architecture}
Our architectural design, inspired by the BEVFusion framework \cite{liu2022bevfusion}, has been meticulously adapted to cater to the specialized demands of joint pruning. As illustrated in Fig. \ref{fig:training_arch}, the architecture seamlessly integrates five pivotal components. Firstly, a sparse encoding backbone processes multi-view camera data, complemented by a counterpart dedicated to LiDAR point cloud inputs. The BEV encoder then synthesizes a cohesive feature map, merging the attributes from both the camera and LiDAR modalities. A perception task head is then incorporated, equipped for either 3D object detection or map segmentation. Parallel to the backbone models, a pruning index predictor is attached, which produces a binary mask facilitating the targeted removal of non-essential raw data inputs.

\textbf{LiDAR Backbone}
For the LiDAR backbone, we adapted the original VoxelNet \cite{zhou2018voxelnet} with a skip mechanism to bypass unnecessary cells in the voxelized 3D space. This skip operation is governed by the binary decision mask rendered by the pruning index predictor, enabling the efficient elimination of superfluous regions from the raw input.

\textbf{Camera Backbone}
Similarly for the camera backbone, we adapted the original DeiT \cite{deit2021hugo} with a drop mechanism to categorize multi-view image into a voxel grid of dimensions \( W \times H \times Z \) through input patchification, and subsequently glean preliminary features. These features then undergo semantic augmentation in the visual attention modules and are finally reshaped by a view transformer implemented by LSS \cite{2020liftsplatshoot} to constitute the BEV feature.

\textbf{Index Predictor and Index Multiplier}
As illustrated in Fig. \ref{fig:training_arch}, the index predictor plays a crucial role in producing a pruning index of dimensions \( W \times H \times Z \), which maps to discrete occupancy positions within a three-dimensional space. The index values are binary, with 0 indicating the exclusion of the corresponding region from the original sensor data, and 1 signifying its inclusion. Architecturally, the predictor comprises several residual blocks, processing front-view images to generate a binary mask as output.

In the inference phase, as depicted in Fig. \ref{fig:testing_arch}, an index multiplier operation is utilized to apply the pruning index onto the voxelized inputs, selectively removing voxels indicated by a 0. This operation involves a straightforward linear algebra procedure that inverts the camera-to-BEV transformation process utilized in LSS \cite{2020liftsplatshoot}, leveraging both the intrinsic and extrinsic parameters of the camera and LiDAR.

\textbf{BEV Encoder}
Following the original design of BEVFusion, we utilize Second \cite{second2018} as the encoder to transcribe the BEV features. However, in this context, the BEV Encoder adopts an additional responsibility, which is to act as a comprehensive feature map, thereby facilitating the supervision of the pruning index predictor's training process.

\textbf{Task Heads}
Our model was subjected to rigorous evaluations on two distinct perception tasks: 3D detection and map segmentation, using the pruned BEV feature map. The detection head leverages the BEV feature as its foundation. Drawing from the CenterPoint \cite{centerpoint2021} method, the prediction targets encompass the position, scale, orientation, and velocity of the entities in the autonomous driving scenarios. For the map segmentation task, the design parallels that of BEVFusion \cite{liu2022bevfusion}, with the prediction target being the segmentation masks representing the various road structures.

\subsection{Multi-Modal Joint Input Pruning}
\textbf{Loss Function}
The joint training of the predictor with the perception model ensures the pruned inputs still contain significant information for the downstream task. The combined loss function \( \mathcal{L}_{total} \) is defined as:
\begin{equation}
        \mathcal{L}_{\text{total}} = \mathcal{L}_{\text{task}} + 
        \alpha \mathcal{L}_{\text{cons}} + 
        \beta \mathcal{L}_{\text{sparse}} + 
        \gamma \mathcal{L}_{\text{penalty}}
\end{equation}
where \( \alpha, \beta, \text{and h} \gamma \) are weights to balance the components. Each item in the total loss is explained as follows:
\begin{enumerate}
    \item \textit{Task Loss:} Standard perception task loss (e.g., segmentation loss, detection loss). Denote as \(\mathcal{L}_{\text{task}}\).
    \item \textit{Consistency Loss:} Measures the similarity between the perception model's outputs and the pruned model's outputs, denoted as \(\mathcal{L}_{\text{cons}}\). We adopted mean squared error (MSE) loss to measure the difference between the original and pruned BEV representation.
    \begin{equation}
        \mathcal{L}_{\text{cons}} = \Vert \text{BEV}_{\text{original}} - \text{BEV}_{\text{pruned}} \Vert^2
    \end{equation}

    \item \textit{Sparsity Loss:} Ensures the predictor mask is sparse enough based on the drop ratio, denoted as \(\mathcal{L}_{\text{sparse}}\). Given the binary mask \( M \) where each element \( M_j \in \{0, 1\} \), the desired ratio \( r \), and total number of elements \( N \), the actual ratio of "0" values in \( M \) is defined as:
    \begin{equation}
        r_{\text{actual}} = \frac{\sum_{j}(1 - M_j)}{N}
    \end{equation}
    The sparsity loss is:
    \begin{equation}
        \mathcal{L}_{\text{sparse}} = (\frac{\sum_{j}(1 - M_j)}{N} - r)^2
    \end{equation}

    \item \textit{Task Penalty Loss:} This loss penalizes the predictor if the mask leads to a significant drop in perception performance, denoted as \( \mathcal{L}_{\text{penalty}} \). Let's consider \( P_{\text{masked}} \) as the perception model's performance with the pruned input and \( P_{\text{original}} \) as its performance without any pruning. Then:
    \begin{equation}
    \mathcal{L}_{\text{penalty}} = \lambda \cdot \max(0, P_{\text{original}} - P_{\text{masked}})
    \end{equation}

    This formulation ensures that \( \mathcal{L}_{\text{mask}} \) is zero if \( P_{\text{masked}} \) surpasses or equals \( P_{\text{original}} \).
\end{enumerate}

\textbf{Training Method}
In the training phase, our objective is to cultivate an efficient predictor capable of discerning the intricacies of the present scene and making judicious decisions regarding regions to omit. To realize this objective, we employ a comparative approach: the predictor is trained by analyzing the performance disparities between a fully optimised anchor model and its pruned counterpart. This is accomplised through a meticulous four-step training strategy.

\begin{enumerate}
    \item Train the perception task model without the predictor using \( \mathcal{L}_{\text{task}} \).
    
    \item With the trained perception model as the anchor, freeze the backbone and task head and start training the predictor. The goal is to make the pruned BEV feature map \(F\) to be as similar as possible to the original, thereby ensuring \( F_{\text{masked}} \) is close to \( F_{\text{original}} \). Train using \( \mathcal{L}_{\text{cons}} \).
    
    \item Train the predictor and perception task model end-to-end together using \( \mathcal{L}_{\text{total}} \).
    
    \item Determine the drop ratio for the predictor, prune the raw input accordingly, and finetune the pruned model using \( \mathcal{L}_{\text{penalty}} \).
\end{enumerate}

\textbf{Inference Method}
In the training phase, the predictor operates in tandem with the primary end-to-end perception task, seamlessly integrated within the data flow. For the inference stage, however, the predictor's role is adjusted to function before the sparse encoding pipeline. As depicted in Fig. \ref{fig:testing_arch}(a), the predictor initially determines a pruning index, a binary decision mask with dimensions \(W \times H \times Z\). This index, in conjunction with an index multiplier as illustrated in Fig. \ref{fig:testing_arch}(c), facilitates the removal of superfluous regions and adjusts the positioning of each voxelized input cell via the input skipping mechanism. After this pruning and realignment step, the refined inputs are forwarded to the sparse encoding backbones (shown in Fig. \ref{fig:testing_arch}(d)), creating a pruned feature map. This pruned map is then utilized by the BEV encoder to produce the final BEV feature map, which is subsequently processed by the task head for output derivation.

This methodology encompasses three pivotal operations: firstly, the deployment of a predictor to establish a pruning mask that identifies less important regions; secondly, the use of an input voxelization module (as demonstrated in Fig. \ref{fig:testing_arch}(b)), which organizes the multi-view images and LiDAR points into a voxel grid matching the dimensions of the pruning mask, a crucial step as it can link the 0 values in the binary mask to those regions in the voxelized input; and thirdly, the application of an index multiplier that performs backward projection (refer to Fig. \ref{fig:mutual_projection}) with sensor parameters to delete the designated redundant regions from the raw sensor inputs before conveying the remainder to the sparse encoding framework

\section{Experiments}
\begin{table*}[t]
\begin{center}
\setlength\tabcolsep{4pt} 
\begin{tabular}{c|c|c|c|c|c| c c c c c}
\hline
Model & Input &  * GFlops $\downarrow$ & * Latency (ms) $\downarrow$ & mAP $\uparrow$ & NDS $\uparrow$ & mATE $\downarrow$ & mASE $\downarrow$ & mAOE $\downarrow$ & mAVE $\downarrow$ & mAAE $\downarrow$ \\
\hline 
DETR3D \cite{detr3d}                & C & 1016.8 & 476.2   & 30.3 & 37.4 & 0.860 & 0.278 & 0.437 & 0.967 & 0.235 \\
PETR \cite{liu2022petr}             & C & 938.5  & 294.1   & 35.7 & 42.1 & 0.710 & 0.270 & 0.490 & 0.885 & 0.224 \\
BEVDepth \cite{li2022bevdepth}      & C & 902.6  & 200.1   & 41.2 & 53.5 & 0.565 & 0.266 & 0.358 & 0.331 & 0.190 \\
BEVFormer \cite{li2022bevformer}    & C & 1303.5 & 588.2   & 41.6 & 47.6 & 0.673 & 0.274 & 0.372 & 0.394 & 0.198 \\
BEVDet4D \cite{huang2022bevdet4d}   & C & 2989.2 & 526.3   & 42.1 & 54.5 & 0.579 & 0.258 & 0.329 & 0.301 & 0.191 \\
CAPE \cite{cape2023xiong}           & C & 1147.7 & 304.9   & 43.9 & 47.9 & 0.683 & 0.267 & 0.427 & 0.814 & 0.197 \\
\hline
$\dagger$ MVP  \cite{mvp2021yin}                            & C+L  & 817.7  & 187.6  & 62.7 & 67.9 & 0.258 & 0.235  & 0.366  & 0.263 & 0.132  \\
$\dagger$ PointAug  \cite{pointaugmenting2021wang}   & C+L  & 939.5  & 263.9  & 64.4 & 67.1 & 0.306 & 0.239  & 0.347  & 0.242 & 0.135  \\
$\dagger$ FUTR3D \cite{futr3d2023chen}                      & C+L  & 2131.3 & 337.4  & 64.6 & 68.1 & 0.258 & 0.240  & 0.333  & 0.232 & 0.127  \\
$\dagger$ TransFusion \cite{transfusion2022bai}             & C+L  & 991.0  & 163.5  & 67.5 & 70.9 & 0.250 & 0.233  & \bf0.308  & \bf0.193 & 0.124  \\
$\dagger$ BEVFusion \cite{liu2022bevfusion}                 & C+L  & 582.3  & 131.0  & 68.6 & 71.3 & 0.255 & 0.242  & 0.340  & 0.219 & \bf0.123  \\
$\dagger$ CMT \cite{cmt2023yan}                             & C+L  & 854.3  & 238.9  & \bf69.1 & \bf72.9 & 0.258 & 0.252  & 0.316  & 0.221 & 0.137  \\
\hline
Ours (Anchor)                                     & C+L & 652.4 & 141.2   & 68.4 & 72.2 & \bf0.242 & \bf0.229 & 0.319 & 0.228 & 0.129 \\
Ours (50\% Pruned)                                  & C+L & \bf423.8 & \bf97.3 & 66.8 & 70.8 & 0.253 & 0.234 & 0.353 & 0.267 & 0.126 \\
\hline
\end{tabular}
\end{center}
\caption{Comparison against SOTA 3D Detection Methods on NuScenes Val Dataset}
\label{table:sota_3d}
\textbf{NOTE}: Performance of the SOTA methods are lifted from the original papers due to the absence of the model weights in the code releases. 
$\dagger$ We re-implemented all C+L methods with open-source code. * We tested GFlops and Latency benchmark with Nvidia 4090 card.
\end{table*}

\begin{table*}[t]
\begin{center}
\setlength\tabcolsep{4pt} 
\begin{tabular}{c|c|c|c|c|c c c c c c}
\hline
Model             & Input & GFlops $\downarrow$ & Latency (ms) $\downarrow$ & Mean IoU $\uparrow$ & Dribable  & Ped. Cross  & Walkway & Stopline & Carpark & Divider  \\
\hline
Image2Map \cite{image2map2022}              & C   & \bf314.4     & 167.4 &    25.0  &    72.6  & 36.3  & 32.4  & -     & 30.5 & -     \\
CVT \cite{cvt2022zhou}                      & C   &    1205.3    & 587.5 &    40.2  &    74.3  & 36.8  & 39.9  & 25.8  & 35.0 & 29.4  \\
LSS \cite{2020liftsplatshoot}               & C   &    742.4     & 371.1 &    44.4  &    75.4  & 38.8  & 46.3  & 30.3  & 39.1 & 36.5  \\
BEVSegFormer \cite{2022bevsegformer}        & C   &    1422.7    & 630.2 &    44.6  &    50.0  & 32.6  & -     & -     & -    & 51.1  \\
Ego3RT \cite{ego3rt2022lu}                  & C   &    1109.6    & 323.6 &    55.5  &    79.6  & 48.3  & 52.0  & -     & 50.3 & 47.5  \\
\hline         
HDMapNet \cite{2021hdmapnet}                & C+L &    952.2     & 264.3 &    44.5  &    56.0  & 31.4  & -     & -     & -    & 46.1 \\
BEVFusion \cite{liu2022bevfusion}           & C+L &    649.6     & 142.9 & \bf62.7  & \bf85.5  & 60.5  & \bf67.6  & \bf52.0  & \bf57.0 & 53.7 \\
\hline   
Ours (Anchor)                               & C+L &    683.8     & 147.7 &    62.2  &    84.9  & \bf61.6  & 67.3  & 51.7  & 56.4 & \bf55.1 \\
Ours (30\% Pruned)                          & C+L &    516.3  & \bf111.7 &    61.4  &    83.3  & 59.2  & 66.3  & 51.0  & 55.6 & 53.9 \\
\hline
\end{tabular}
\end{center}
\caption{Comparison with SOTA Map Segmentation Methods on NuScenes Val Dataset}
\label{table:sota_map}
\end{table*}

\subsection{Implementation Details}
Our implementation closely mirrors the configurations presented in BEVFusion \cite{liu2022bevfusion}, ensuring consistency in processing protocols. The dataset employed for our experiments is NuScenes. For 3D detection, our evaluation criteria encompass mAP and NDS (NuScenes Detection Score), while segmentation performance is gauged using the mIoU (Intersection over Union) metric. The original dimensions of the NuScenes dataset are $1600 \times 900$, which we rescale to an input resolution of $1408\times512$. The data augmentation pipeline comprises random transformations such as flipping, scaling, cropping, and rotation. To address the inherent class imbalance, we integrate a copy-paste mechanism, complemented by the Class-Balanced-Grouping-and-Sampling (CBGS) \cite{cbgs2019} technique during the training phase, akin to the approach employed by CenterPoint \cite{centerpoint2021}. In the testing phase, scaling adjustments are made to align the image resolution with the model's input requirements.

For the camera modality, we deploy DeiT \cite{deit2021hugo} as the primary backbone, employing LSS \cite{2020liftsplatshoot} for projecting 2D imagery into the BEV space. The LiDAR modality leverages the VoxelNet \cite{zhou2018voxelnet} backbone. Consistent with BEVFusion's methodology, we stack BEV features derived from both camera and LiDAR, subsequently integrating them via a convolutional network. The defined LiDAR input span and voxel range are (-51.2, -51.2, -8) to (51.2, 51.2, 8) and (0.1, 0.1, 0.2), respectively. The predictor is implemented with Resnet34 \cite{resnet2015he} with an output dimension of 128x128x16. The binarization threshold for the decision mask is set at 0.5. Each unit within this output correlates to the native sensor space, facilitated by intrinsic and extrinsic sensor parameters. LiDAR is designated as the primary sensor for pruning superfluous zones. Indices in the multi-view image domain that are misaligned are rounded to synchronize with the LiDAR domain. 

The training regimen employs the AdamW optimizer, set with a learning rate of 1e-4 and a weight-decay of 1e-2. As proposed in \cite{liu2022bevfusion}, the learning rate is modulated to linearly ascend to 1e-3 during the initial 40\% of the training regimen and subsequently diminish to zero. During the perception task training stage, we maintain a batch size of 4 across eight A100 GPU cards, training across 24 epochs for each experiment iteration and archiving the optimal model configuration for subsequent evaluations. During the subsequent three independent finetuning stages, we train for 24 epochs each and keep the best for next stage.

\subsection{Benchmark Results}
Having elucidated our experimental setups, this section details the comparative assessment of our joint pruning method against contemporary SOTA methods. We scrutinize its efficacy across two distinct perception tasks pivotal for autonomous driving.

\textbf{3D Detection}
Table \ref{table:sota_3d} provides a comprehensive overview of various models' performance. At the zenith of this list, CMT \cite{cmt2023yan} registers the highest mAP and NDS at 69.1 and 72.9 respectively. Pitted against this benchmark is our anchor model, devoid of any pruning interventions, attaining commendable results. While our mAP and NDS are slightly trailling those of CMT, the differential is insignificant. In a side-by-side comparison with our baseline, BEVFusion, our anchor model exhibits a slightly subdued mAP but trumps in the NDS. This affirms the model's effectiveness as a robust reference for the pruning paradigm. With a pruning ratio of 50\%, our model achieves significant efficiency, registering a complexity reduction of 35\% (from 652.4 to 423.8) and a 31\% contraction in inference latency (to 97.3 ms from 141.2 ms). Importantly, these gains were accompanied by a trivial mAP and NDS drop of 1.6 and 1.4 points respectively, thus underscoring the success of our joint pruning approach.

\textbf{Map Segmentation}
Pertaining to the map segmentation domain, Table \ref{table:sota_map} shows analogous trends as Table \ref{table:sota_3d}. BEVFusion emerges as the premier model for this task. Nevertheless, our anchor model, with no pruning, mirrors BEVFusion's performance with only a slight deviation. With a 30\% input pruning ratio, our anchor model registers a minimal decline of 0.8 points in mIoU. This negligible decrease is judiciously counterbalanced by the significant enhancement in model simplicity offered by our methodology, reinforcing the versatility and robustness of our approach.

\section{Ablation Study}
\subsection{Effect of Different Pruning Methods}
In this investigative study, we aim to evaluate the efficacy of our novel Multimodal-Joint-Pruning (MJP) approach against four recognized pruning methodologies: magnitude-based weight pruning \cite{weightpruning2017zhuang}, filter pruning \cite{filterpruning2017hao}, structural pruning \cite{depgraph23gongfan}, and a method akin to ours, learning-based pruning \cite{prune_kong2022spvit}, with a particular focus on the 3D detection and map segmentation tasks. To enhance the study's efficiency and articulation, we reduce the standard training epoch duration to 12, subsequently reporting the average performance across the specified tasks.

\begin{table}[h]
\begin{center}
\begin{tabular}{c|c|c|c|c}
\hline
Method                                              & mAP       & mIoU  & GFlops    & Latency (ms)\\
\hline
Weight \cite{weightpruning2017zhuang}               & 27.3      & 22.4  & 549.2     & 125.5 \\
Filter \cite{filterpruning2017hao}                  & 13.9      & 11.0  & \bf155.7  & \bf49.4 \\
Structural \cite{depgraph23gongfan}                 & 42.5      & 34.1  & 412.0     & 97.2 \\
Learning \cite{prune_kong2022spvit}                 & 43.8      & 34.6  & 486.7     & 111.4 \\
\hline
Ours (MJP)                                          & \bf45.1 & \bf35.8  & 423.4    & 95.3 \\
\hline
\end{tabular}
\end{center}
\caption{Comparisons of Different Pruning Methods on NuScenes Val Dataset}
\label{table:deletion_method}
\textbf{NOTE} * MJP: Our Multimodal-Joint-Pruning method. 
\end{table}

Results in Table \ref{table:deletion_method} show that our MJP method outperforms in mAP and mIoU with scores of 45.1 and 35.8, respectively, indicating enhanced model accuracy and segmentation performance. Despite Filter Pruning's computational gains, it falls short in precision. Weight Pruning, while less demanding computationally, compromises performance. The Structural and Learning-based Pruning methods do incur a tradeoff between efficiency and accuracy; however, neither attains the performance achieved by MJP. 

In summary, our investigation affirms the predominance of the Multimodal-Joint-Pruning method, underscoring its effectiveness and efficiency against a range of conventional pruning strategies.

\begin{figure}[t]
\begin{center}
\includegraphics[scale=0.85, width=0.9\linewidth]{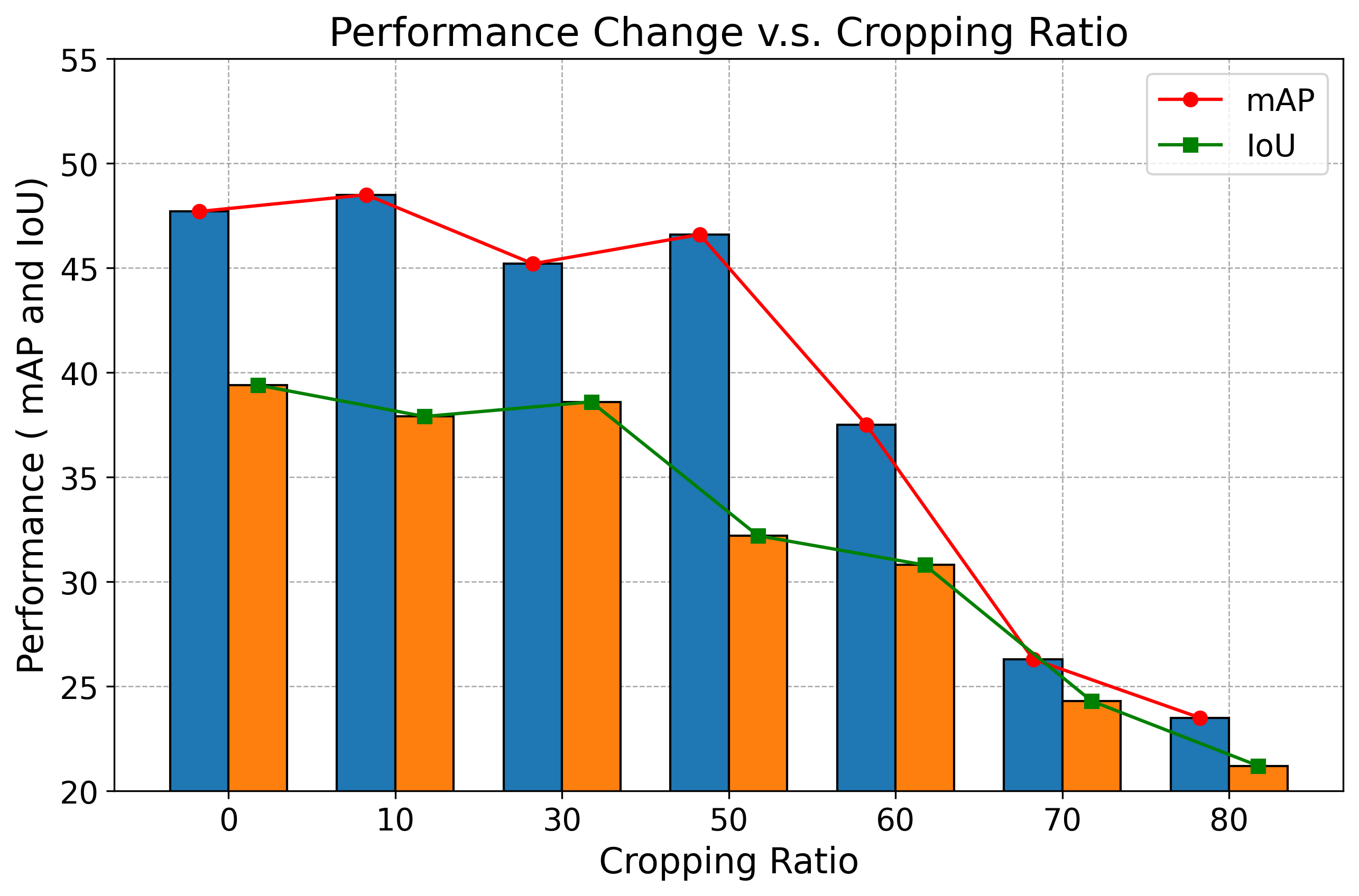}
\end{center}
\caption{Comparison of Performance Change w.r.t. Cropping Ratio. Models are trained for 12 epochs instead of 24 for fast verification purposes.}
\label{fig:cropping}
\end{figure}

\subsection{Performance Variations with Cropping Ratio}
The impact of the cropping ratio on model performance is a core focus of our study. To evaluate this relationship, we conducted a series of experiments with our proposed method. We trained the model over 12 epochs for these specific tests, half the usual 24. As depicted in Fig. \ref{fig:cropping}, both mAP and IoU metrics for 3D detection and map segmentation tasks were considered.

For 3D detection, the model's performance generally declined with increased pruning. A marginal performance decline was observed when the pruning ratios were up to 50\%,  however, a significant drop happened when the ratio was increased to 60\%. Interestingly, a 10\% pruning ratio led to a slight performance improvement. We speculate this might act similarly to a dropout mechanism, forcing the model to prioritize more crucial data.

In contrast, map segmentation displayed a greater sensitivity to pruning. With pruning ratios of 10\% and 30\%, the decline in performance was subtle, remaining within 2 percentage points of the unpruned model. However, at a 50\% ratio, the performance decreased substantially by 7\%. This can be attributed to the segmentation task's necessity to predict complete scene elements, making it inherently more sensitive to pruning than the 3D detection. Despite these challenges, our method proved its merit, successfully reducing input sizes without significant performance degradation.

\begin{figure}[t]
\begin{center}
\includegraphics[scale=0.85, width=0.9\linewidth]{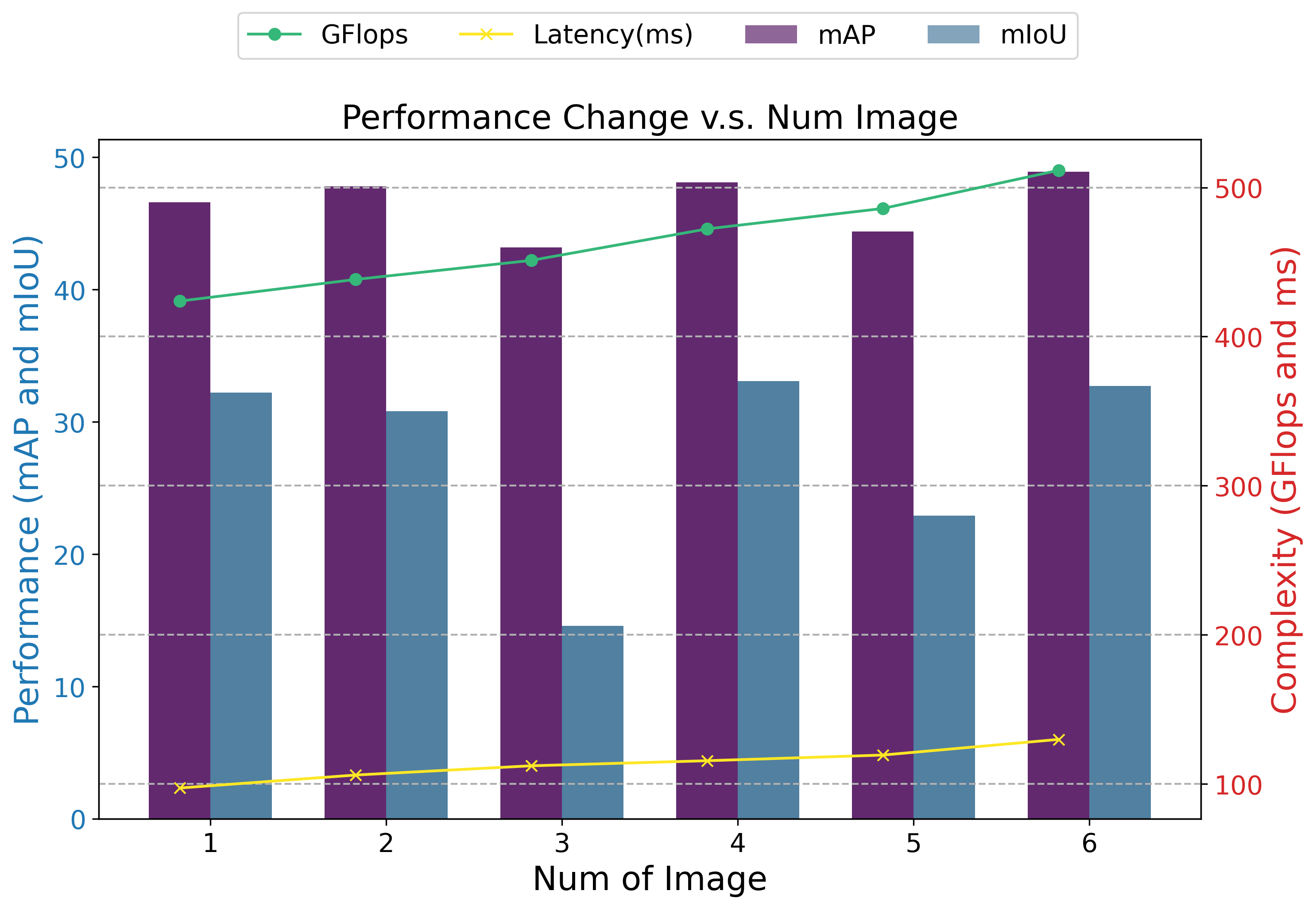}
\end{center}
\caption{Comparison of Performance Change w.r.t. Number of images used to generate pruning mask. Pruning ratio is set to be 50\% for fair comparison. Models are trained for 12 epochs instead of 24 for fast verification purpose.}
\label{fig:viewstudy}
\end{figure}

\subsection{Performance Variations with Different Numbers of Views}
This study aims to ascertain the ideal number of images necessary for effective pruning mask generation, analyzing the impact of incremental image count on model complexity. Our methodology commenced with a singular front view, progressively incorporating additional perspectives up to a total of six (including front, back, front left, front right, back left and back right views) to ensure a holistic representation of all the possible views. The findings from this systematic exploration are depicted in Fig. \ref{fig:viewstudy}.

Counterintuitively, the results reveal that increasing the count of images from the surrounding camera angles does not consistently correlate with improved model performance. Specifically, the mAP yields a limited increase of up to 5\% (rising from 46.6 to 48.9) with the inclusion of up to six images. Similarly, the mIoU experiences a slight elevation (from 32.2 to 33.1) with four and six images. As expected, increasing image count is accompanied by a substantial rise in computational demand, marked by a 20\% increase in GFlops (from 423.8 to 511.5) and a 33\% hike in latency (from 97.3 to 123.9 ms). 

This phenomenon may be attributed to the fact that vehicles predominantly move forward, and the integration of a temporal fusion module, which retains information from previous frames, ensures that the front view sufficiently captures the most critical regions for effective model performance. Although augmenting the number of views might seem beneficial for enhancing the overall pipeline performance, this increase in computational load compromises the model's efficiency, contravening our initial objective. Given that our primary goal is to develop an efficient perception model that offers performance comparable to state-of-the-art methods, opting for only the front view in generating the pruning mask represents a pragmatic yet effective choice.

\section{Conclusion}
In this paper, we propose a novel domain-specific pruning method for efficient perception tasks in BEV space. Our method leverages the BEV feature map as a shared anchor to jointly remove redundant sensor inputs before passing them into the backbone model and thus effectively reduce the model complexity. Our experiments show that our method can significantly reduce the model complexity with marginal performance tradeoff, leading to faster and more resource-efficient inference. For future work, we would like to explore a more efficient end-to-end training pipeline and other alternative methods for addressing the complexity of BEV perception methods.





\newpage

\bibliographystyle{IEEETran}
\bibliography{IEEEexample,myref}

\begin{thebibliography}{10}
\providecommand{\url}[1]{#1}
\csname url@rmstyle\endcsname
\providecommand{\newblock}{\relax}
\providecommand{\bibinfo}[2]{#2}
\providecommand\BIBentrySTDinterwordspacing{\spaceskip=0pt\relax}
\providecommand\BIBentryALTinterwordstretchfactor{4}
\providecommand\BIBentryALTinterwordspacing{\spaceskip=\fontdimen2\font plus
\BIBentryALTinterwordstretchfactor\fontdimen3\font minus \fontdimen4\font\relax}
\providecommand\BIBforeignlanguage[2]{{%
\expandafter\ifx\csname l@#1\endcsname\relax
\typeout{** WARNING: IEEEtran.bst: No hyphenation pattern has been}%
\typeout{** loaded for the language `#1'. Using the pattern for}%
\typeout{** the default language instead.}%
\else
\language=\csname l@#1\endcsname
\fi
#2}}

\bibitem{li2022bevformer}
Z.~L. et~al, ``Bevformer: Learning bird's-eye-view representation from multi-camera images via spatiotemporal transformers,'' in \emph{European Conference on Computer Vision (ECCV)}, vol. 13669.\hskip 1em plus 0.5em minus 0.4em\relax Springer, 2022, pp. 1--18.

\bibitem{2022beverse}
Y.~Zhang, Z.~Zhu, W.~Zheng, J.~Huang, G.~Huang, J.~Zhou, and J.~Lu, ``Beverse: Unified perception and prediction in birds-eye-view for vision-centric autonomous driving,'' \emph{arXiv}, vol. abs/2205.09743, 2022.

\bibitem{xie2022m2bev}
E.~Xie, Z.~Yu, D.~Zhou, J.~Philion, A.~Anandkumar, S.~Fidler, P.~Luo, and J.~M. {\'{A}}lvarez, ``M\({}^{\mbox{2}}\)bev: Multi-camera joint 3d detection and segmentation with unified birds-eye view representation,'' \emph{arXiv}, vol. abs/2204.05088, 2022.

\bibitem{huang2022bevdet4d}
J.~Huang and G.~Huang, ``Bevdet4d: Exploit temporal cues in multi-camera 3d object detection,'' \emph{arxiv}, vol. abs/2203.17054, 2022.

\bibitem{li2022bevdepth}
Y.~L. et~al, ``Bevdepth: Acquisition of reliable depth for multi-view 3d object detection,'' \emph{arxiv}, vol. abs/2206.10092, 2022.

\bibitem{huang2021bevdet}
J.~Huang, G.~Huang, Z.~Zhu, Y.~Yun, and D.~Du, ``Bevdet: High-performance multi-camera 3d object detection in bird-eye-view,'' \emph{arXiv}, 2021.

\bibitem{liu2022bevfusion}
Z.~Liu, H.~Tang, A.~Amini, X.~Yang, H.~Mao, D.~Rus, and S.~Han, ``Bevfusion: Multi-task multi-sensor fusion with unified bird's-eye view representation,'' in \emph{IEEE International Conference on Robotics and Automation (ICRA)}, 2023.

\bibitem{harley2022simplebev}
A.~W. Harley, Z.~Fang, J.~Li, R.~Ambrus, and K.~Fragkiadaki, ``Simple-bev: What really matters for multi-sensor {BEV} perception?'' in \emph{{IEEE} International Conference on Robotics and Automation, {ICRA}}.\hskip 1em plus 0.5em minus 0.4em\relax {IEEE}, 2023, pp. 2759--2765.

\bibitem{futr3d2023chen}
X.~Chen, T.~Zhang, Y.~Wang, Y.~Wang, and H.~Zhao, ``{FUTR3D:} {A} unified sensor fusion framework for 3d detection,'' in \emph{{IEEE/CVF} Conference on Computer Vision and Pattern Recognition Workshop, {CVPRW}}.\hskip 1em plus 0.5em minus 0.4em\relax {IEEE}, 2023, pp. 172--181.

\bibitem{cmt2023yan}
J.~Yan, Y.~Liu, J.~Sun, F.~Jia, S.~Li, T.~Wang, and X.~Zhang, ``Cross modal transformer: Towards fast and robust 3d object detection,'' \emph{arxiv}, vol. abs/2301.01283, 2023.

\bibitem{2021hdmapnet}
Q.~Li, Y.~Wang, Y.~Wang, and H.~Zhao, ``Hdmapnet: An online {HD} map construction and evaluation framework,'' in \emph{2022 International Conference on Robotics and Automation, {ICRA}}.\hskip 1em plus 0.5em minus 0.4em\relax {IEEE}, 2022, pp. 4628--4634.

\bibitem{2022bevsegformer}
L.~Peng, Z.~Chen, Z.~Fu, P.~Liang, and E.~Cheng, ``Bevsegformer: Bird's eye view semantic segmentation from arbitrary camera rigs,'' in \emph{{IEEE/CVF} Winter Conference on Applications of Computer Vision, {WACV} 2023, Waikoloa, HI, USA, January 2-7, 2023}.\hskip 1em plus 0.5em minus 0.4em\relax {IEEE}, 2023, pp. 5924--5932.

\bibitem{li2022vectormapnet}
Y.~Liu, T.~Yuan, Y.~Wang, Y.~Wang, and H.~Zhao, ``Vectormapnet: End-to-end vectorized {HD} map learning,'' in \emph{International Conference on Machine Learning, {ICML}}, ser. Proceedings of Machine Learning Research, A.~Krause, E.~Brunskill, K.~Cho, B.~Engelhardt, S.~Sabato, and J.~Scarlett, Eds., vol. 202.\hskip 1em plus 0.5em minus 0.4em\relax {PMLR}, 2023, pp. 22\,352--22\,369.

\bibitem{maptr2023liao}
B.~Liao, S.~Chen, X.~Wang, T.~Cheng, Q.~Zhang, W.~Liu, and C.~Huang, ``Maptr: Structured modeling and learning for online vectorized {HD} map construction,'' in \emph{International Conference on Learning Representations, {ICLR}}, 2023.

\bibitem{weightpruning2017zhuang}
Z.~Liu, J.~Li, Z.~Shen, G.~Huang, S.~Yan, and C.~Zhang, ``Learning efficient convolutional networks through network slimming,'' in \emph{{IEEE/CVF} International Conference on Computer Vision, {ICCV}}.\hskip 1em plus 0.5em minus 0.4em\relax {IEEE}, 2017.

\bibitem{filterpruning2017hao}
H.~Li, A.~Kadav, I.~Durdanovic, H.~Samet, and H.~P. Graf, ``Pruning filters for efficient convnets,'' in \emph{International Conference on Learning Representations, {ICLR}}.\hskip 1em plus 0.5em minus 0.4em\relax OpenReview.net, 2017.

\bibitem{depgraph23gongfan}
\BIBentryALTinterwordspacing
G.~Fang, X.~Ma, M.~Song, M.~B. Mi, and X.~Wang, ``Depgraph: Towards any structural pruning,'' in \emph{{IEEE/CVF} Conference on Computer Vision and Pattern Recognition, {CVPR} 2023, Vancouver, BC, Canada, June 17-24, 2023}.\hskip 1em plus 0.5em minus 0.4em\relax {IEEE}, 2023, pp. 16\,091--16\,101. [Online]. Available: \url{https://doi.org/10.1109/CVPR52729.2023.01544}
\BIBentrySTDinterwordspacing

\bibitem{prune_astar2021}
A.~Parnami, R.~Singh, and T.~Joshi, ``Pruning attention heads of transformer models using a* search: {A} novel approach to compress big {NLP} architectures,'' \emph{arXiv}, vol. abs/2110.15225, 2021.

\bibitem{prune_kong2022spvit}
Z.~Kong, P.~Dong, X.~Ma, X.~Meng, W.~Niu, M.~Sun, X.~Shen, G.~Yuan, B.~Ren, H.~Tang, M.~Qin, and Y.~Wang, ``Spvit: Enabling faster vision transformers via latency-aware soft token pruning,'' in \emph{European Conference on Computer Vision (ECCV)}, 2022, pp. 620--640.

\bibitem{prune_ltp2021}
S.~Kim, S.~Shen, D.~Thorsley, A.~Gholami, W.~Kwon, J.~Hassoun, and K.~Keutzer, ``Learned token pruning for transformers,'' in \emph{Conference on Knowledge Discovery and Data Mining, {KDD}}, A.~Zhang and H.~Rangwala, Eds.\hskip 1em plus 0.5em minus 0.4em\relax {ACM}, 2022, pp. 784--794.

\bibitem{prune_toconv2021}
H.~He, J.~Liu, Z.~Pan, J.~Cai, J.~Zhang, D.~Tao, and B.~Zhuang, ``Pruning self-attentions into convolutional layers in single path,'' \emph{arxiv}, vol. abs/2111.11802, 2021.

\bibitem{zhou2018voxelnet}
Y.~Zhou and O.~Tuzel, ``Voxelnet: End-to-end learning for point cloud based 3d object detection,'' in \emph{{IEEE} Conference on Computer Vision and Pattern Recognition, {CVPR}}, 2018, pp. 4490--4499.

\bibitem{deit2021hugo}
H.~Touvron, M.~Cord, M.~Douze, F.~Massa, A.~Sablayrolles, and H.~J{\'{e}}gou, ``Training data-efficient image transformers {\&} distillation through attention,'' in \emph{International Conference on Machine Learning, {ICML}}, ser. Proceedings of Machine Learning Research, vol. 139.\hskip 1em plus 0.5em minus 0.4em\relax {PMLR}, 2021, pp. 10\,347--10\,357.

\bibitem{2020liftsplatshoot}
J.~Philion and S.~Fidler, ``Lift, splat, shoot: Encoding images from arbitrary camera rigs by implicitly unprojecting to 3d,'' in \emph{European Conference on Computer Vision (ECCV)}, 2020.

\bibitem{second2018}
Y.~Yan, Y.~Mao, and B.~Li, ``{SECOND:} sparsely embedded convolutional detection,'' \emph{Sensors}, vol.~18, no.~10, p. 3337, 2018.

\bibitem{centerpoint2021}
T.~Yin, X.~Zhou, and P.~Kr{\"{a}}henb{\"{u}}hl, ``Center-based 3d object detection and tracking,'' in \emph{{IEEE} Conference on Computer Vision and Pattern Recognition, {CVPR}}.\hskip 1em plus 0.5em minus 0.4em\relax Computer Vision Foundation / {IEEE}, 2021, pp. 11\,784--11\,793.

\bibitem{detr3d}
Y.~Wang, V.~Guizilini, T.~Zhang, Y.~Wang, H.~Zhao, and J.~M. Solomon, ``Detr3d: 3d object detection from multi-view images via 3d-to-2d queries,'' in \emph{The Conference on Robot Learning ({CoRL})}, 2021.

\bibitem{liu2022petr}
Y.~L. et~al, ``{PETR:} position embedding transformation for multi-view 3d object detection,'' in \emph{European Conference on Computer Vision (ECCV)}, vol. 13687.\hskip 1em plus 0.5em minus 0.4em\relax Springer, 2022, pp. 531--548.

\bibitem{cape2023xiong}
K.~X. et~al, ``{CAPE:} camera view position embedding for multi-view 3d object detection,'' \emph{arxiv}, vol. abs/2303.10209, 2023.

\bibitem{mvp2021yin}
T.~Yin, X.~Zhou, and P.~Kr{\"{a}}henb{\"{u}}hl, ``Multimodal virtual point 3d detection,'' in \emph{Advances in Neural Information Processing Systems (NeurIPS)}, 2021, pp. 16\,494--16\,507.

\bibitem{pointaugmenting2021wang}
C.~Wang, C.~Ma, M.~Zhu, and X.~Yang, ``Pointaugmenting: Cross-modal augmentation for 3d object detection,'' in \emph{{IEEE} Conference on Computer Vision and Pattern Recognition, {CVPR}}.\hskip 1em plus 0.5em minus 0.4em\relax {IEEE}, 2021, pp. 11\,794--11\,803.

\bibitem{transfusion2022bai}
X.~Bai, Z.~Hu, X.~Zhu, Q.~Huang, Y.~Chen, H.~Fu, and C.~Tai, ``Transfusion: Robust lidar-camera fusion for 3d object detection with transformers,'' in \emph{{IEEE/CVF} Conference on Computer Vision and Pattern Recognition, {CVPR}}.\hskip 1em plus 0.5em minus 0.4em\relax {IEEE}, 2022, pp. 1080--1089.

\bibitem{image2map2022}
A.~Saha, O.~Mendez, C.~Russell, and R.~Bowden, ``Translating images into maps,'' in \emph{International Conference on Robotics and Automation, {ICRA}}.\hskip 1em plus 0.5em minus 0.4em\relax {IEEE}, 2022, pp. 9200--9206.

\bibitem{cvt2022zhou}
B.~Zhou and P.~Kr{\"{a}}henb{\"{u}}hl, ``Cross-view transformers for real-time map-view semantic segmentation,'' in \emph{{IEEE} Conference on Computer Vision and Pattern Recognition, {CVPR}}.\hskip 1em plus 0.5em minus 0.4em\relax {IEEE}, 2022, pp. 13\,750--13\,759.

\bibitem{ego3rt2022lu}
J.~Lu, Z.~Zhou, X.~Zhu, H.~Xu, and L.~Zhang, ``Learning ego 3d representation as ray tracing,'' in \emph{{IEEE/CVF} European Conference on Computer Vision, {ECCV}}, ser. Lecture Notes in Computer Science, vol. 13686.\hskip 1em plus 0.5em minus 0.4em\relax Springer, 2022, pp. 129--144.

\bibitem{cbgs2019}
B.~Zhu, Z.~Jiang, X.~Zhou, Z.~Li, and G.~Yu, ``Class-balanced grouping and sampling for point cloud 3d object detection,'' \emph{arxiv}, vol. abs/1908.09492, 2019.

\bibitem{resnet2015he}
K.~H. et~al, ``Deep residual learning for image recognition,'' in \emph{{IEEE} Conference on Computer Vision and Pattern Recognition, {CVPR}}.\hskip 1em plus 0.5em minus 0.4em\relax {IEEE} Computer Society, 2016, pp. 770--778.

\end{thebibliography}

\end{document}